\documentclass[11pt]{article}

\usepackage[preprint]{acl}

\usepackage{times}
\usepackage{latexsym}

\usepackage[T1]{fontenc}

\usepackage[utf8]{inputenc}

\usepackage{microtype}

\usepackage{inconsolata}

\usepackage{graphicx}
\usepackage{multirow}

\usepackage{booktabs}
\usepackage{tabularx}

\usepackage[most]{tcolorbox}
\usepackage{listings}
\usepackage{enumitem}

\newtcolorbox{appendixbox}[1]{
    enhanced,
    breakable,
    title={#1},
    colback=gray!7,
    colframe=gray!35,
    colbacktitle=gray!60,
    coltitle=white,
    fonttitle=\bfseries,
    boxrule=0.5pt,
    arc=2pt,
    outer arc=2pt,
    left=2mm,
    right=2mm,
    top=1.5mm,
    bottom=1.5mm,
    before skip=6pt,
    after skip=6pt
}

\newtcblisting{promptbox}[1]{
    enhanced,
    breakable,
    listing only,
    listing engine=listings,
    title={#1},
    colback=gray!7,
    colframe=gray!35,
    colbacktitle=gray!60,
    coltitle=white,
    fonttitle=\bfseries,
    boxrule=0.5pt,
    arc=2pt,
    outer arc=2pt,
    left=2mm,
    right=2mm,
    top=1.5mm,
    bottom=1.5mm,
    before skip=6pt,
    after skip=6pt,
    listing options={
        basicstyle=\ttfamily\footnotesize,
        breaklines=true,
        breakatwhitespace=true,
        columns=fullflexible,
        keepspaces=true,
        showstringspaces=false,
        tabsize=2
    }
}

\usepackage{amsmath}
\usepackage{amssymb}
\usepackage{booktabs}
\usepackage{tabularx}
\usepackage{array}

\title{ReRef-3D: A Benchmark for Spatial Referring Expression-Guided 3D Object Placement}

\author{
   Mary Lynn Martin \,\,\,
   Yifei Zhang \,\,\,
   Martha Palmer \,\,\,
   Maria Leonor Pacheco \\
   University of Colorado Boulder \\
   \texttt{\{mary.martin, yifei.zhang, martha.palmer, maria.pacheco\}@colorado.edu} \\
}

\begin{document}
\maketitle
\begin{abstract}
We introduce ReRef-3D, a benchmark for language-guided single-object placement in controlled 3D tabletop scenes. It contains 33,826 instructions across 998 CLEVR-derived scenes, spanning 16 placement types and direct, one-hop, and two-hop references. Given that an instruction defines a region of acceptable placements rather than one coordinate, predictions are evaluated by relation satisfaction and physical validity. Each instruction also includes a verified natural rephrasing. We evaluate LLaVA-3D, 3D-LLM, 3D-LLaVA, LL3DA, and PlaceIt3D, with PlaceIt3D evaluated under both point-based and region-based supervision. Region-supervised PlaceIt3D achieves the highest placement accuracy at about 80\% and LLaVA-3D reaches about 70\%. Performance varies by relation type, but \emph{nearest} and \emph{aligned} remain the most difficult. Rephrasing has little effect on performance overall.
\end{abstract}

\section{Introduction}
\label{sec:intro}

Spatial language shapes how people describe actions in the physical world. Most 3D vision-language benchmarks evaluate understanding of a fixed scene through grounding, question answering, captioning, or text-based planning \citep{hong20233dllminjecting3dworld, chen2024grounded3dllmreferenttokens, zhu2025llava3dsimpleeffectivepathway}. Embodied manipulation systems generate actions, but are generally evaluated through task-specific policies and output spaces \citep{shridhar2021cliportpathwaysroboticmanipulation, shridhar2022perceiveractormultitasktransformerrobotic, jiang2023vimageneralrobotmanipulation, huang2023voxposercomposable3dvalue}. 

Recent placement methods use realistic scans and 3D assets \citep{abdelreheem2025placeit3dlanguageguidedobjectplacement, huang2025fireplacegeometricrefinementsllm} or operate in robot manipulation settings \citep{kim2024lingospacelanguageconditionedincrementalgrounding, yuan2024robopointvisionlanguagemodelspatial}, where perception and commonsense expectations are entangled with spatial reasoning. Less attention has gone to whether 3D language models can turn a spatial instruction into a verifiable scene change when spatial reasoning is isolated from other factors that could influence model behavior.

\begin{figure}[t]
    \centering
    \includegraphics[width=\columnwidth]{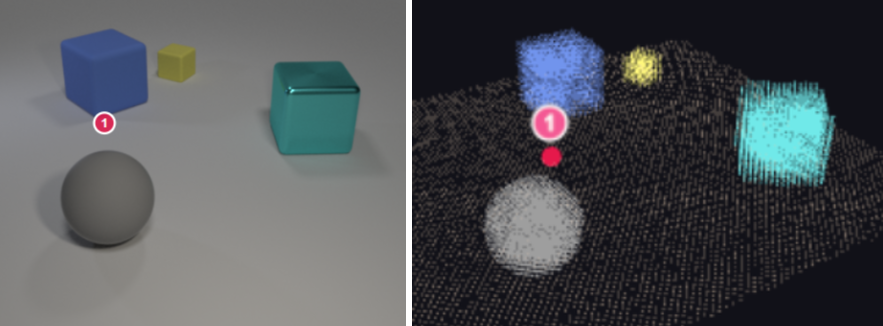}
    \caption{Paired target location (\textcolor{red}{1}) for ``Place the gray ball next to the block that is to the left of the yellow block that is behind the cyan block.'' In 2D and 3D. 
    }
    \label{fig:placement_example}
\end{figure}

We introduce ReRef-3D, a benchmark that targets this gap by requiring models to resolve a self-contained relational instruction into a new position in a 3D tabletop scene. In Figure~\ref{fig:placement_example}, a model must follow a two-hop reference chain to identify the anchor block, then predict a collision-free position next to it on the table. 
Unlike scene generation methods that arrange many objects \citep{feng2023layoutgptcompositionalvisualplanning, yang2024holodecklanguageguidedgeneration, lin2024instructsceneinstructiondriven3dindoor}, ReRef-3D places a single object to focus on spatial reference resolution and placement reasoning, sitting between referring-expression grounding and broader rearrangement.

In Figure~\ref{fig:placement_example}, 
many positions place the gray ball next to the intended block, and the marked location is only one of them. This means that distance from a single annotated point is an incomplete measure of success. We therefore evaluate the scene after the object has been moved, checking whether the instruction’s relations are satisfied and whether the placement is physically valid. A relation-satisfying placement can still be invalid if it leaves the tabletop or collides with another object. Instructions may combine multiple constraints or use relations such as \textit{between}, \textit{nearest}, and \textit{farthest}, which depend on several objects or the arrangement of the scene \citep{liu2021structformerlearningspatialstructure}. Related evaluators score the spatial and physical properties of whole generated scenes~\citep{tam2026sceneevalevaluatingsemanticcoherence, sengupta2026scenecriticsymbolicevaluator3d}, while we check whether a single edit satisfies the relations in its instruction. 

ReRef-3D uses simple, easily recognized objects so errors are more likely to reflect failures in reference resolution or placement reasoning than visual perception. The controlled scenes also reduce ambiguity from object function and affordances, which influence how people interpret spatial relations \citep{Coventry1994SpatialPO, krishnaswamy2025affordanceembeddings}, so the benchmark focuses on explicit spatial semantics and does not seek to capture the full range of pragmatic spatial language. Known scene geometry allows relations and clearance to be recomputed after each placement. Each example includes a structured 3D scene, a self-contained instruction, symbolic spatial constraints, and a verified goal location, which is only one valid solution. 

We specifically address four research questions: \textbf{(RQ1)} Can existing 3D language models learn language-guided continuous placement? \textbf{(RQ2)} Which spatial relations and reference structures remain difficult? \textbf{(RQ3)} How do failure patterns differ across model families? \textbf{(RQ4)} How does linguistic variation affect placement performance?

Our main contributions are:

\begin{itemize}[nosep,leftmargin=*]
    \item \textbf{Procedural generation with verification.}
    We retain placements that satisfy the requested relations, workspace constraints, and collision checks.
    \item \textbf{Evaluation beyond a single reference coordinate.}
    We convert coordinate and region outputs to a common representation and score post-move relation satisfaction and physical validity.
    \item \textbf{Two phrasings of every instruction.}
    We provide each expression in templated form and as a rephrasing designed to sound more natural, with human verification on a sampled subset.
\end{itemize}

\section{Background and Related Work}

\paragraph{Spatial referring expressions and 3D grounding.}

RefCOCO, RefCOCO+, and RefCOCOg are widely used natural-image referring expression datasets \citep{yu2016modelingcontextreferringexpressions, mao2016generationcomprehensionunambiguousobject}. CLEVR-Ref+ uses controlled synthetic scenes with functional programs and compositional expressions for diagnostic evaluation \citep{liu2019clevrrefdiagnosingvisualreasoning}. In 3D, ReferIt3D and ScanRefer focus on a single referent, while ScanEnts3D grounds other objects mentioned in the expression and Multi3DRefer has multiple referents \citep{achlioptas2020referit_3d, chen2020scanrefer3dobjectlocalization, abdelreheem2023scanents3dexploitingphraseto3dobjectcorrespondences, zhang2023multi3drefergroundingtextdescription}. Relation-aware methods like ViL3DRel, EDA, and VIGOR model anchor objects and referential order \citep{chen2022languageconditionedspatialrelation, Wu_2023, wang2024learningvisualgroundinggenerative}. 

General-purpose 3D language models extend these capabilities. 3D-LLM, LL3DA, LLaVA-3D, and 3D-LLaVA use varied scene representations and vision-language interfaces \citep{hong20233dllminjecting3dworld, chen2023ll3davisualinteractiveinstruction, deng20253dllavageneralist3dlmms, zhu2025llava3dsimpleeffectivepathway}. We evaluate these models to investigate how spatial understanding transfers to continuous placement.



\paragraph{Language-guided object placement and rearrangement.}

Language-conditioned manipulation systems map instructions to object poses, actions, or spatial affordances. CLIPort, PerAct, VIMA, and VoxPoser use pick-and-place actions, discretized poses, multimodal prompts, or spatial value maps \citep{shridhar2021cliportpathwaysroboticmanipulation, shridhar2022perceiveractormultitasktransformerrobotic, jiang2023vimageneralrobotmanipulation, huang2023voxposercomposable3dvalue}. 



More closely related to our setting, LINGO-SPACE grounds compositional expressions as distributions over a target space and RoboPoint predicts image-space affordance points using auto-generated supervision \citep{kim2024lingospacelanguageconditionedincrementalgrounding, yuan2024robopointvisionlanguagemodelspatial}. RoboSpatial uses paired 2D and 3D supervision to evaluate spatial affordances and manipulation \citep{song2026robospatialteachingspatialunderstanding}. PlaceIt3D predicts valid placement regions for an object provided separately from the scene \citep{abdelreheem2025placeit3dlanguageguidedobjectplacement}. ReRef-3D describes the moved object and anchors defining its destination within one self-contained instruction. FirePlace also studies language-guided 3D placement with richer geometric and functional constraints \citep{huang2025fireplacegeometricrefinementsllm}. ReRef-3D controls the setting to emphasize compositional spatial reasoning and placement.

\paragraph{3D scene generation, editing, rearrangement, and evaluation.}

StructFormer, StructDiffusion, and LGMCTS study object rearrangement with pose generation, structure prediction, and constraint based planning \citep{liu2021structformerlearningspatialstructure, liu2023structdiffusionlanguageguidedcreationphysicallyvalid, chang2024lgmctslanguageguidedmontecarlotree}. Systems like LayoutGPT, Holodeck, and InstructScene generate multi-object layouts or complete 3D environments \citep{feng2023layoutgptcompositionalvisualplanning, yang2024holodecklanguageguidedgeneration, lin2024instructsceneinstructiondriven3dindoor}. ScanEdit allows language-guided editing of real 3D scans, including object orientation \citep{boudjoghra2025scanedithierarchicallyguidedfunctional3d}. ReRef-3D narrows the problem into a single-object placement task under explicit spatial constraints.

SceneEval and SceneCritic evaluate generated 3D scenes with spatial, semantic, and geometric criteria \citep{tam2026sceneevalevaluatingsemanticcoherence, sengupta2026scenecriticsymbolicevaluator3d}. Our evaluation applies similar concepts to assess whether each instructed placement satisfies the requested relation and is physically valid.

\section{Dataset Generation}

\subsection{Task Definition}
\label{sec:task-def}

We represent a scene as $S = (\mathcal{O},\mathcal{R},\mathcal{D})$, where $\mathcal{O}=\{o_1,\ldots,o_n\}$ is the set of objects, $\mathcal{R}$ is the relational structure, and $\mathcal{D}$ contains camera-relative direction vectors used to define spatial relations.

Each object $o_i$ has semantic attributes (color, shape, size, and material) and geometric information: a three-dimensional center $\mathbf{p}_i=(x_i,y_i,z_i)$ and a ground-plane position $\mathbf{q}_i=(x_i,y_i)$. We use \emph{tabletop} to refer to the bounded, planar support surface within which valid placements occur. Placement is represented on this plane, with the workspace denoted by $\mathcal{T}\subset\mathbb{R}^2$. The primary prediction is the target's new ground-plane position, while its vertical coordinate is determined by its radius. The scene and spatial relations are three-dimensional, but the target moves only along the tabletop plane.

Given a scene $S$ and a placement instruction $e$, the task is to predict a new position for the target specified in $e$. Let $t_e$ be the target object, and $\mathcal{C}(e)=\{c_1,\ldots,c_K\}$ denote the instruction's spatial constraints. Moving $t_e$ to a position $\mathbf{q}$ produces the updated scene $S'=\operatorname{Move}(S,t_e,\mathbf{q})$, and the feasible placement set is

\begin{equation}\small
\mathcal{F}(S,e)
=
\left\{
\mathbf{q}\in\mathcal{T}
\;\middle|\;
\begin{aligned}
&\bigwedge_{k=1}^{K} c_k(S',t_e)=1,\\
&\operatorname{PhysValid}(S',t_e)=1
\end{aligned}
\right\},
\end{equation}

The annotated goal $\mathbf{q}^{*}$ is one verified member of $\mathcal{F}(S,e)$. It serves as a point-based supervision target, but it is not the only correct placement. Models can also be trained against the valid placement region, as described in Section \ref{sec:eval-setup}.


\subsection{CLEVR-Ref+ Starting Point}

ReRef-3D is constructed from CLEVR-derived scenes and builds on the object vocabulary, scene geometry, camera-relative relations, and description machinery of CLEVR-Ref+ \citep{liu2019clevrrefdiagnosingvisualreasoning}. The source scenes contain objects that vary in color, shape, size, material, and position.

To support language-guided placement, we introduce new templates and functional operators that describe the target and specify its desired placement in one expression, as:
\begin{quote}
``Place the small red block to the left of the green cylinder.''
\end{quote}
This identifies both the object to move and the anchor that defines its intended location.

Our templates extend beyond pairwise directions to all placement families in Section~\ref{sec:spatial-rel}, combined with direct, one-hop, two-hop, and ordinal anchor descriptions. The placement constraints, goal generation, and post-move verification are new, while CLEVR-Ref+ provides the scenes, object vocabulary, and program-execution framework. Splits are assigned by scene, yielding 23{,}334 training expressions over 699 scenes, 5{,}248 validation expressions over 151 scenes, and 5{,}244 test expressions over 148 scenes: 33{,}826 expressions across 998 scenes. 

\subsection{Generation Strategy}

Generation operates over scene-object pairs. Each object is considered a potential target, and we generate a controlled number of valid expressions for that object. This reduces target-level imbalance. The procedure also controls the relative frequency of placement types, so no family dominates the dataset. The final expression must refer to the target unambiguously.

\subsection{3D Scene Representation}

Each scene contains object-centered semantic and geometric information. Objects are assigned one of two radii, $r_i=0.35$ if $o_i$ is small and $r_i=0.70$ if it is large, and the vertical center of a moved target is set to $z_t=r_t$, keeping it on the tabletop.

The scene also provides the camera-relative unit vectors
$\hat{\mathbf{d}}_{\mathrm{left}}$,
$\hat{\mathbf{d}}_{\mathrm{right}}$,
$\hat{\mathbf{d}}_{\mathrm{front}}$, and
$\hat{\mathbf{d}}_{\mathrm{behind}}$, with
($\hat{\mathbf{d}}_{\mathrm{left}}=-\hat{\mathbf{d}}_{\mathrm{right}}$ and
$\hat{\mathbf{d}}_{\mathrm{front}}=-\hat{\mathbf{d}}_{\mathrm{behind}}$).

The relational representation records directional and between relations, and the nearest-to-farthest ordering of objects relative to each reference object. These values are recomputed after every proposed move. Image-plane coordinates and depth estimate whether the target would be hidden behind another object.

\begin{figure*}[t]
    \centering
    \includegraphics[width=0.8\linewidth]{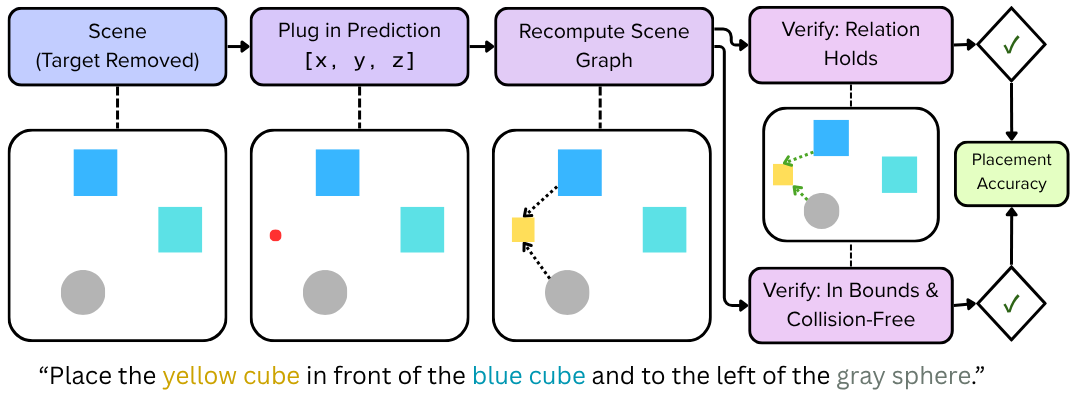}
    \caption{Placement verification used during dataset generation and evaluation. The proposed placement is shown in \textcolor{red}{red}. After the placement, the scene graph is recomputed and checked against the instruction. \textcolor{green}{Green edges} indicate verified relations.}
    \label{fig:scene-verify}
\end{figure*}

\subsection{Spatial Relation Modeling}
\label{sec:spatial-rel}

\paragraph{Placement constraint families.}

ReRef-3D contains the following placement families:

\begin{itemize}[nosep, leftmargin=*]
    \item \textbf{Pairwise directional:} the target must be left, right, in front of, or behind one anchor.
    
    \item \textbf{Aligned:} the target must satisfy a directional relation within a narrow angular cone and remain close to the anchor.
    
    \item \textbf{Between:} the target must lie between two anchors.
    
    \item \textbf{Nearest and farthest:} the target must occupy an extreme position in the distance ordering around an anchor; nearest also requires adjacency.
    
    \item \textbf{Multi-relation:} the target must satisfy two directional constraints simultaneously.
    
    \item \textbf{Ordinally anchored:} the anchor is identified by its ordinal position among same-shape objects before a directional relation is applied.

    \item \textbf{Table-region:} the target must be placed in a named center, edge, or corner region in the camera frame.
\end{itemize}

Placement geometry is independent of reference complexity. Direct, one-hop, two-hop, and ordinal expressions may resolve to the same anchor and use the same placement geometry; they differ only in the language required to identify that anchor.

Throughout, subscripts $t$ and $a$ denote the target and anchor, and $\mathbf{q}_t$ and $\mathbf{q}_a$ their ground-plane positions. 

\paragraph{Directional relations.}


For target $t$, anchor $a$, and directional relation $r$, let $\hat{\mathbf{d}}_r^\perp$ be a unit vector perpendicular to $\hat{\mathbf{d}}_r$. We decompose the target-anchor displacement into an axial component $u_r=(\mathbf{q}_t-\mathbf{q}_a)^\top\hat{\mathbf{d}}_r$ and a lateral component $\ell_r=(\mathbf{q}_t-\mathbf{q}_a)^\top\hat{\mathbf{d}}_r^\perp$. We define $c_r(t,a)=1$ when $u_r>\tau$ and $|\ell_r|\leq\rho u_r$, where $\tau=0.2$ and $\rho=1.0$, corresponding to a $45^\circ$ half-angle.

This cone requires movement in the requested direction to exceed the lateral offset, and the positive axial margin $\tau$ excludes positions nearly coincident with the anchor. CLEVR's native relations instead use loose half-planes of the form $(\mathbf{q}_t-\mathbf{q}_a)^\top\hat{\mathbf{d}}_r>0$, which can admit far off-axis placements. This cone is used for goal generation; prediction evaluation uses the relaxed relation tolerance described in Section~\ref{sec:eval-metrics}.

\paragraph{Aligned relations.}

Expressions containing terms such as \textit{directly} require a stricter version of the corresponding directional relation. Let $g=0.3$ denote the minimum gap between objects, and $d_{\mathrm{clear}}=r_t+r_a+g$ the collision-free clearance. Using the same axial and lateral components, an aligned relation holds when
\begin{equation}\small
\begin{aligned}
u_r &> \tau, \\
|\ell_r| &\leq
\max\!\left(\varepsilon_0,\tan\theta\,u_r\right), \\
u_r &\leq d_{\mathrm{clear}}+\beta.
\end{aligned}
\end{equation}
where $\theta=7^\circ$, $\varepsilon_0=0.12$, and $\beta=0.6$.

The angular tolerance applies a consistent alignment criterion across distances. The bound $u_r\leq d_{\mathrm{clear}}+\beta$ prevents aligned placements from being too far from the anchor.

\paragraph{Between relations.}


A between relation uses two anchors, $a_1$ and $a_2$. Let $\mathbf{s}=\mathbf{q}_{a_2}-\mathbf{q}_{a_1}$ be the segment joining them, and let $\hat{\mathbf{s}}^\perp$ be a perpendicular unit vector. Candidate target positions take the form $\mathbf{q}_t=\mathbf{q}_{a_1}+\lambda\mathbf{s}+w\hat{\mathbf{s}}^\perp$, where $\lambda\in[0.2,0.8]$ and $w\in[-0.3\|\mathbf{s}\|_2,\,0.3\|\mathbf{s}\|_2]$.

This band is used only for proposal generation; after the move, the target must satisfy the between relation with the required anchor pair.

\paragraph{Nearest and farthest relations.}

A nearest instruction, such as ``next to'' or ``closest to,'' requires the target to become the closest object to the anchor while remaining adjacent to it:
\begin{equation}\small
\begin{aligned}
\|\mathbf{q}_t-\mathbf{q}_a\|_2
&<
\min_{j\notin\{t,a\}}
\|\mathbf{q}_j-\mathbf{q}_a\|_2,
\\
\|\mathbf{q}_t-\mathbf{q}_a\|_2
&\leq
d_{\mathrm{clear}}+\beta_{\mathrm{adj}}.
\end{aligned}
\end{equation}
where $\beta_{\mathrm{adj}}=0.6$. The adjacency bound prevents a distant object in a sparse scene from satisfying ``next to'' because it is the nearest object.



A farthest instruction requires $\|\mathbf{q}_t-\mathbf{q}_a\|_2 > \max_{j\notin\{t,a\}}\|\mathbf{q}_j-\mathbf{q}_a\|_2$.

\paragraph{Table-region placement.}

Table-region instructions are the only anchor-free placement family and specify one of nine absolute regions: the center, four edges, or four corners.

Because the CLEVR camera views the square table at roughly $41^\circ$, these regions are defined in the camera frame. Candidate positions are projected onto the scene's front and left unit vectors, so ``front'' refers toward the camera and ``left'' to camera-left. Region membership is determined using fixed fractional thresholds based on the table projection, with the same thresholds used during generation and verification. Sampling is biased toward the requested edge or corner so that accepted goals are not near the central object cluster.

\paragraph{Multi-relation placement.}

For a multi-relation instruction containing $(r_1,a_1)$ and $(r_2,a_2)$, a valid placement must satisfy both constraints, $c_{r_1}(t,a_1)\land c_{r_2}(t,a_2)$. Candidates are sampled from the overlap of the two relation-specific proposal regions and verified in the updated scene.

\paragraph{Goal generation.}

Each accepted example includes a verified goal point $\mathbf{q}^\star$, generated in three stages. First, candidates are drawn from a relation-specific proposal region: a directional or aligned cone, a band around an anchor segment, a nearest-neighbor annulus, a far field, or a named table region. Second, each candidate is filtered for workspace boundaries and collisions. It must satisfy $|x|,|y|\leq 3.0$ and remain at least $r_t+r_j+g$ from every other object $j$. Finally, the move is applied, the scene graph is recomputed, and the constraints are checked, as shown in Figure~\ref{fig:scene-verify}. Candidates that fail verification are discarded, and constraints with no feasible goal produce no example.

Because the first candidate to pass every check is retained, the procedure produces varied draws from the verified region. Deterministic candidates are used only when the feasible region is too narrow for sampling. Each item uses a fixed scene- and placement-specific random seed for reproducibility.

\subsection{Instruction Generation}

Each placement instruction combines an unambiguous target description, one or more anchor descriptions, an action verb, and one or more placement predicates. It is paired with a functional symbolic representation that resolves the relevant objects before applying the terminal placement constraint.

Before goal generation, the symbolic representation is executed against the source scene to confirm that the attribute filters, relational descriptions, ordinal operations, and uniqueness constraints resolve the intended objects. The procedure also rejects no-op instructions. If $C_e(S,t_e)$ denotes the requested condition, an accepted example must satisfy $C_e(S,t_e)=0$.

\subsection{Improving Expression Naturalness}
\label{sec:expression-naturalization}

\paragraph{Candidate generation.} Template generation provides precise control over spatial meaning but can produce mechanically phrased instructions. We therefore use Qwen3-8B \citep{yang2025qwen3} to generate four rewrites under explicit semantic constraints, followed by verification and selection. The generation prompt preserves every object, attribute, and relation while allowing linguistic variation, balancing semantic fidelity and paraphrase diversity \citep{bandel-etal-2022-quality}. Full prompts and generation settings are provided in Appendix~\ref{app:candidate-generation-settings}.

\paragraph{Candidate verification and selection.} We use Gemini 3.5 Flash to verify the four rewrites and select the best candidate. Gemini filters out candidates that fail to preserve the intended semantics. Valid candidates are scored for grammatical correctness, fluency and natural wording, and clarity, then ranked. If none are valid, the original expression is retained. Full prompts are provided in Appendix~\ref{app:verification-selection-settings}. 

\paragraph{Human evaluation.} Two annotators independently evaluated four rewrites for each of 100 sampled expressions, with coverage across all 16 placement types. Both annotators judged all 400 candidates to be semantically valid, and Gemini agreed for 366 (91.5\%). Most disagreements involved lexical realizations of distance-based relations. The annotators selected the same best candidate for 40 of the 100 expressions ($\kappa=0.19$), which likely reflects subjective preference among several semantically valid and similarly natural rewrites. On those 40, Gemini matched the human consensus in 25 cases (62.5\%, $\kappa=0.48$). Full annotation procedures and analysis are provided in Appendix~\ref{app:naturalness-evaluation-details}.

\begin{figure*}[t]
    \centering
    \includegraphics[width=0.9\linewidth, height=7cm, keepaspectratio=false]{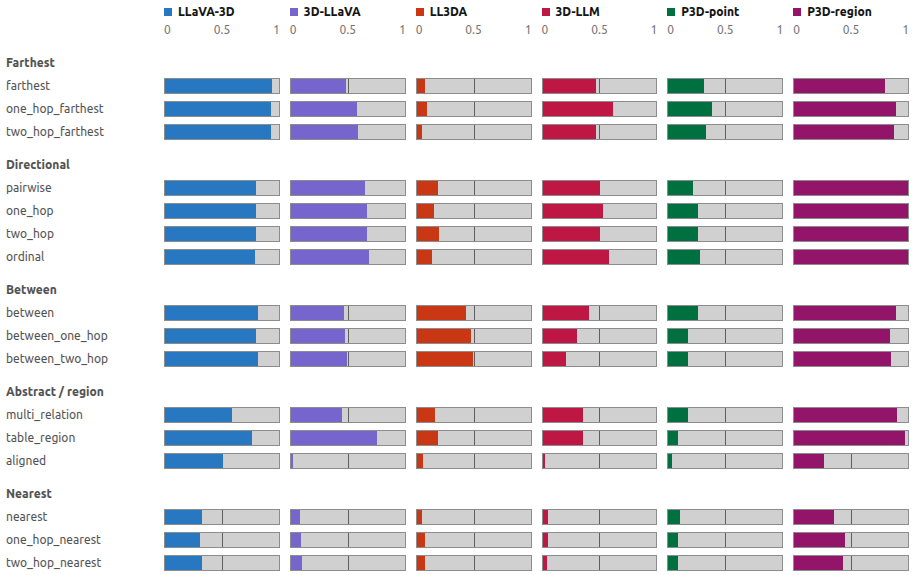}
    \caption{ Per-model placement accuracy by template type for templated instructions. One row per placement type, one column per model. Bar length is the score on a shared 0–1 track; the tick marks 0.5.}
    \label{fig:per-type-template}
\end{figure*}

\section{Evaluation}

\subsection{Evaluation Setup}
\label{sec:eval-setup}

\begin{table*}[t]
\centering
\small
\setlength{\tabcolsep}{5pt}
\begin{tabular}{lccccccc}
\toprule
Model
& \shortstack{LLaVA\\family}
& \shortstack{Generalist\\3D}
& \shortstack{Direct\\3D}
& \shortstack{Lifted\\multi-view}
& \shortstack{Placement-\\specific}
& \shortstack{Point\\sup.}
& \shortstack{Region\\sup.} \\
\midrule
\textbf{LLaVA-3D}
& $\checkmark$ &  &  & $\checkmark$ &  & $\checkmark$ &  \\

\textbf{3D-LLaVA}
& $\checkmark$ &  & $\checkmark$ &  &  & $\checkmark$ &  \\

\textbf{3D-LLM}
&  & $\checkmark$ &  & $\checkmark$ &  & $\checkmark$ &  \\

\textbf{LL3DA}
&  & $\checkmark$ & $\checkmark$ &  &  & $\checkmark$ &  \\

\textbf{PlaceIt3D (point)}
&  &  & $\checkmark$ &  & $\checkmark$ & $\checkmark$ &  \\

\textbf{PlaceIt3D (region)}
&  &  & $\checkmark$ &  & $\checkmark$ &  & $\checkmark$ \\
\bottomrule
\end{tabular}
\caption{Comparison of the evaluated model configurations.}
\label{tab:model-comparison}
\end{table*}

For our evaluation, we use five models and six configurations: LLaVA-3D, 3D-LLaVA, 3D-LLM, LL3DA, and PlaceIt3D with point-based and region-based supervision \citep{zhu2025llava3dsimpleeffectivepathway, deng20253dllavageneralist3dlmms, hong20233dllminjecting3dworld, chen2023ll3davisualinteractiveinstruction, abdelreheem2025placeit3dlanguageguidedobjectplacement}. All models are fine-tuned separately on the template and rephrased instructions and evaluated on both, giving matched and cross-phrasing conditions for the same scenes. 

Models are scored on the same examples with the same placement constraints and procedure for evaluation. Because they differ in scene representation and placement generation, cross-model results are behavioral comparisons, not controlled architectural ablations. Table~\ref{tab:model-comparison} summarizes the configurations.

Both PlaceIt3D configurations use the model's region output. The point-supervised version is trained around the sampled goal location, while the region-supervised version uses a set of placements that satisfy the constraints of the instruction. During evaluation, the predicted region is reduced to a superpoint location and scored identically to coordinate predictions. Because region supervision also changes parts of the training recipe, we refer to it as a region-supervised recipe.

\paragraph{Zero-shot performance.} Zero-shot performance on templated instructions is near the floor for most models, with several models failing to emit parseable placements; full results and discussion are provided in Appendix~\ref{app:zero-shot}.

\subsection{Metrics}
\label{sec:eval-metrics}

The primary metric is placement accuracy. Predictions are considered correct if they satisfy the requested spatial relations and are physically valid. Physical validity requires the prediction to stay within the tabletop workspace and avoid collisions. Relation satisfaction and physical validity are also reported separately.

\paragraph{Relation tolerance.} Goal generation uses more conservative relation thresholds than prediction evaluation. The reported results use an evaluation tolerance of $1.5\times$, relaxing the spatial thresholds while leaving the \textit{between} criterion unchanged. For directional relations, the allowable cone is widened from 45$^{\circ}$ to about 63$^{\circ}$. Tabletop boundary and collision checks are not relaxed.  

\paragraph{Human placement evaluation.} To determine whether our placement criteria aligned with human judgments, three annotators selected locations for the same 100 instructions. These instructions were sampled to cover all 16 placement types. For this task, annotators viewed the instruction, rendered scene, and an interactive 3D representation. They then selected a placement location for the target object. The size of the target object was shown at placement time so annotators could view whether their placement would fit at the selected location or cause a collision. Instructions noted that they should interpret instructions from the original camera view. Two annotators produced valid placements for the same 97 instructions, while the third produced valid placements for 93. All three failed the evaluation criteria on three instructions, including two table-region corner placements and one multi-relation placement. Four additional instructions produced split judgments, three of which were table-region placements. The full procedures and analysis are provided in Appendix~\ref{app:placement-location-eval}.

\section{Results and Analysis}

\begin{table}[t]
\centering
\small
\setlength{\tabcolsep}{4pt}
\begin{tabular}{llccc}
\toprule
Model & Inst.
& \shortstack{Placement\\acc.}
& \shortstack{Relation\\sat.}
& \shortstack{Physical\\valid.} \\
\midrule
\multirow{2}{*}{LLaVA-3D}
    & T & 69.0 & \textbf{95.3} & 72.6 \\
    & R & 70.2 & \textbf{95.1} & 74.0 \\
\midrule
\multirow{2}{*}{3D-LLaVA}
    & T & 46.8 & 70.2 & 63.4 \\
    & R & 48.2 & 71.4 & 64.3 \\
\midrule
\multirow{2}{*}{3D-LLM}
    & T & 35.9 & 51.2 & 63.2 \\
    & R & 34.9 & 49.9 & 62.0 \\
\midrule
\multirow{2}{*}{LL3DA}
    & T & 15.7 & 70.1 & 25.0 \\
    & R & 18.0 & 70.3 & 26.8 \\
\midrule
\multirow{2}{*}{PlaceIt3D (point)}
    & T & 19.7 & 36.9 & 54.6 \\
    & R & 25.4 & 38.1 & 67.7 \\
\midrule
\multirow{2}{*}{PlaceIt3D (region)}
    & T & \textbf{79.7} & 80.3 & \textbf{98.4} \\
    & R & \textbf{78.8} & 79.7 & \textbf{97.8} \\
\bottomrule
\end{tabular}
\caption{Matched results (\%). T denotes templated and R denotes rephrased instructions.}
\label{tab:main-results}
\end{table}


\paragraph{RQ1: Can existing 3D language models learn language-guided continuous placement?} Yes, 3D language models can learn language-guided continuous placement with varying performance across models and supervision strategies. Table~\ref{tab:main-results} summarizes the matched results. Region-supervised PlaceIt3D performs best, reaching about 80\% placement accuracy on both instruction sets. LLaVA-3D follows with about 70\% and the highest relation satisfaction at over 95\%. 3D-LLaVA achieved 47-48\% and 3D-LLM is around 35\%. LL3DA and point-supervised PlaceIt3D have lower placement accuracy. 

Separating relation satisfaction and physical validity clarifies these differences. About a quarter of LLaVA-3D's predictions fail the physical-validity check, despite satisfying the requested relation. LL3DA's results show a larger gap; relation satisfaction is around 70\%, but only 25-27\% of the predictions are physically valid. Nearly all of region-supervised PlaceIt3D's predictions are physically valid, leaving relation satisfaction as its main source of error. The effect of region-aware training is discussed under RQ3.


\begin{table}[t]
\centering
\scriptsize
\setlength{\tabcolsep}{3pt}
\begin{tabular}{lccccc}
\toprule
Model
& Near
& Far
& Dir.
& Betw.
& Other \\
\midrule
LLaVA-3D
& 31.8 & 93.6 & 79.9 & 80.9 & 60.7 \\
3D-LLaVA
& 8.9 & 55.6 & 66.3 & 47.7 & 36.9 \\
3D-LLM
& 4.3 & 52.6 & 52.6 & 30.2 & 23.0 \\
LL3DA
& 6.5 & 7.3 & 17.5 & 47.1 & 12.9 \\
PlaceIt3D (pt.)
& 9.3 & 34.2 & 24.2 & 19.7 & 9.7 \\
PlaceIt3D (reg.)
& \textbf{41.1} & 85.8 & \textbf{100.0} & \textbf{86.3} & \textbf{68.0} \\
\bottomrule
\end{tabular}
\caption{Placement accuracy (\%) by relation group for templated matched runs. ``Other'' groups aligned, table-region, and multi-relation placements. Full template-level results appear in Figure~\ref{fig:per-type-template}}.
\label{tab:relation-groups}
\end{table}

\paragraph{RQ2: Which spatial relations and reference structures remain difficult?}

\paragraph{Relation families.} Table~\ref{tab:relation-groups} shows the placement accuracy by relation family for templated matched runs and Figure~\ref{fig:per-type-template} illustrates a more detailed breakdown by template. The most obvious difficulty is \textit{nearest}. Even the best performing configurations reach only 41.1\% for PlaceIt3D-region and 31.8\% for LLaVA-3D. The remaining models score lower than 10\%. \textit{Aligned} is also difficult; LLaVA-3D and PlaceIt3D-region score 51.3\% and 25.8\%, respectively. The accuracy for the remaining models is below 6\%.

\begin{table}[t]
\centering
\small
\setlength{\tabcolsep}{4pt}
\begin{tabular}{lccc}
\toprule
Model & Direct & One-hop & Two-hop \\
\midrule
LLaVA-3D
& 73.1 & 66.2 & 68.0 \\
3D-LLaVA
& 51.0 & 41.7 & 44.6 \\
3D-LLM
& 40.8 & 35.0 & 30.4 \\
LL3DA
& 16.8 & 15.9 & 17.1 \\
PlaceIt3D (pt.)
& 21.0 & 21.4 & 20.8 \\
PlaceIt3D (reg.)
& 84.7 & 76.1 & 77.0 \\
\bottomrule
\end{tabular}
\caption{Placement accuracy (\%) by reference depth for templated matched runs.}
\label{tab:reference-depth}
\end{table}

\paragraph{Reference depth.} Table~\ref{tab:reference-depth} compares direct, one-hop, and two-hop instructions within the relation families that contain depth progression (see Figure~\ref{fig:per-type-template} for details). LLaVA-3D is fairly stable as depth increases. Its score is 73.1\% for direct references and decreases to 66.2\% for one-hop expressions, and 68.0\% for two-hop references. Depth has the clearest effect on 3D-LLM, dropping from 40.8\% to 35\% and then 30.4\%. Region-supervised PlaceIt3D drops about eight points from direct to one-hop and then stays stable. 

Reference depth does matter, but the same pattern is not present for every model. We deduce that relation geometry and model placement behavior are greater sources of variation.  

\paragraph{RQ3: How do failure patterns differ across model families?} Across the evaluated models, failures present differently. The main limitation for LLaVA-3D is physical validity, after the relation is already satisfied. 3D-LLM and point-supervised PlaceIt3D are limited the most by relation satisfaction. LL3DA shows the strongest mismatch, as it often satisfies a relation while producing an invalid placement. Only region-supervised PlaceIt3D is high for both relation satisfaction and physical validity.

\paragraph{PlaceIt3D: point vs. region supervision.} Region-supervised PlaceIt3D substantially outperforms its point-supervised counterpart, increasing placement accuracy from 19.7\% to 79.7\%. This evidence supports the use of the valid placement region. Because the region configuration also changes the loss, encoder training, and query mechanism, the gain should be attributed to the region-aware training recipe as a whole.

\paragraph{Model-family comparisons.} Within the LLaVA family, LLaVA-3D outperforms 3D-LLaVA on placement accuracy and relation satisfaction. Because their encoders and training procedures are different, this gap cannot be attributed to scene representation. The generalist 3D models trade off relation satisfaction and physical validity, with LL3DA favoring the former and 3D-LLM the latter.

\paragraph{RQ4: How does linguistic variation affect placement performance?} To answer this, we compare templated instructions with the rephrasings designed to sound more natural. Table~\ref{tab:phrasing-cross} shows performance across all combinations of training and evaluation phrasing. For most models, phrasing has very little effect, with differences of about two percentage points or less between the matched templated and rephrased conditions. However, point-supervised PlaceIt3D improves by 5.7 points when trained and evaluated on rephrased instructions. For the same trained model, switching the phrasing of the evaluation instructions changes placement accuracy by at most 1.5 points across all models. When trained on the rephrased set, LL3DA and point-supervised PlaceIt3D improve by 2.3 and 5.7 points, respectively. Given these results, the phrasing distinction does not appear to be a notable source of difficulty in ReRef-3D. Nevertheless, these results do not imply that the models evaluated here are insensitive to language in general.


\begin{table}[t]
\centering
\small
\setlength{\tabcolsep}{4pt}
\begin{tabular}{lcccc}
\toprule
Model
& T$\rightarrow$T
& T$\rightarrow$R
& R$\rightarrow$T
& R$\rightarrow$R \\
\midrule
LLaVA-3D
& 69.0 & 67.5 & 69.8 & 70.2 \\
3D-LLaVA
& 46.8 & 46.9 & 48.4 & 48.2 \\
3D-LLM
& 35.9 & 35.3 & 35.0 & 34.9 \\
LL3DA
& 15.7 & 15.7 & 18.0 & 18.0 \\
PlaceIt3D (pt.)
& 19.7 & 19.7 & 25.4 & 25.4 \\
PlaceIt3D (reg.)
& 79.7 & 79.3 & 79.0 & 78.8 \\
\bottomrule
\end{tabular}
\caption{Placement accuracy (\%) across train/evaluation phrasing combinations. The first symbol denotes training phrasing and the second evaluation phrasing; T denotes templated and R rephrased.}
\label{tab:phrasing-cross}
\end{table}

\section{Conclusion and Future Directions}

ReRef-3D was created to examine how 3D language models turn compositional spatial instructions into continuous placements. The results show that success depends on both satisfying spatial constraints and producing physically valid locations, with different models exhibiting distinct failure modes. For PlaceIt3D, region-aware supervision substantially improves placement performance. Our plans for future work include extending ReRef-3D to richer geometry, more complex sets of spatial relations, longer reference structures, and more realistic environments. Our human annotation effort is ongoing, and we plan to expand this evaluation in future versions of the benchmark. \textbf{Upon acceptance, we will release the code and data publicly on GitHub.}

\section*{Limitations}

ReRef-3D focuses on a controlled form of spatial placement in which one object is moved on a planar tabletop and its vertical position is fixed by object geometry. Orientation, stacking, alternative support surfaces, and multi-object rearrangement are outside the current scope. This leaves additional forms of 3D manipulation for future work.

The scenes are synthetic and use simple objects under a fixed camera setup. This isolates geometric spatial relations and reference complexity. Real environments may introduce additional effects from object function, affordances, and pragmatic intent.

The model comparison is behavioral, not architectural, since systems differ in backbone, scene representation, pretraining, adaptation, and output format. We also use one fine-tuning configuration per condition.

The PlaceIt3D point-versus-region comparison should be interpreted as a comparison of training recipes, since the region configuration changes more than the supervision target. The general-purpose coordinate models also remain point-supervised even though an instruction may admit multiple valid placements.


\bibliography{custom}

\appendix

\section{Model and Implementation Details}
\label{app:impl}

We fine-tune six model configurations using identical scene splits and instructions. Four predict goal coordinates, while two PlaceIt3D variants predict placement masks.

\subsection{Model Architectures and Adaptation}

Table~\ref{tab:impl-summary} summarizes each model's architecture, scene representation, output, and adaptation strategy.

\begin{table*}[t]
\centering
\small
\setlength{\tabcolsep}{4pt}
\begin{tabular}{p{2.4cm}p{2.1cm}p{4.1cm}p{2.5cm}p{4cm}}
\toprule
Model & Backbone & Scene Representation & Output & Adaptation \\
\midrule
LLaVA-3D
    & Vicuna-7B
    & Posed RGB-D view repeated to 5 frames (1152 patch tokens)
    & Text coordinates
    & LoRA ($r=64$, $\alpha=16$); backbone frozen \\

3D-LLaVA
    & LLaVA-1.5-7B
    & Occupancy point cloud ($\sim$1024 superpoints, 100 point tokens)
    & Text coordinates
    & LoRA ($r=32$, $\alpha=64$); trainable alignment and segmentation projections \\

LL3DA
    & OPT-1.3B
    & Point cloud encoded by Vote2Cap-DETR and Q-Former
    & Discretized \texttt{<loc>} tokens
    & Q-Former trained; detector and LLM frozen \\

3D-LLM
    & Flan-T5-XL
    & Precomputed lifted features ($N \times 1408$) and voxel indices
    & Text coordinates
    & Q-Former and Flan-T5 trained \\

PlaceIt3D-point
    & Flan-T5-XL
    & Point cloud ($\sim$1024 superpoints) and asset embedding
    & Placement mask, rotation bucket, anchor mask
    & Point encoder and T5 body frozen \\

PlaceIt3D-region
    & Flan-T5-XL
    & Same as above
    & Same as above
    & Point encoder unfrozen; T5 body frozen \\
\bottomrule
\end{tabular}
\caption{Model architectures and fine-tuning strategies. Both PlaceIt3D variants use the PlaceWizard architecture.}
\label{tab:impl-summary}
\end{table*}

LLaVA-3D, 3D-LLaVA, and LL3DA are initialized from LLaVA-3D-7B, LLaVA-1.5-7B, and LL3DA generalist checkpoints, respectively. 3D-LLM uses the BLIP-2 SAM Flan-T5-XL checkpoint, and both PlaceIt3D variants use PlaceWizard. The 3D-LLaVA OST tower remains frozen, while LL3DA uses color, height, and zero-valued normals as point features.

\paragraph{Supervision.}
The four coordinate models use the recorded goal position, represented as text coordinates or discretized \texttt{<loc>} tokens for LL3DA.

PlaceIt3D-point predicts a ground-point disc around the goal using BCE and Dice losses, with the placement mask queried at \texttt{[SEG]}. PlaceIt3D-region instead predicts a softmax distribution over the constraint-valid superpoints, using the mean question-token hidden state as its query. The region variant also unfreezes the point encoder, applies a mask bias of $-4$, and uses a three-layer attention warmup. Both variants predict rotation buckets and anchor masks.

\subsection{Training Configuration}

\begin{table*}[t]
\centering
\small
\setlength{\tabcolsep}{4pt}
\begin{tabular}{lccccccc}
\toprule
Model & Epochs & Batch & Peak LR & Schedule & Warmup & Weight Decay & Precision \\
\midrule
LLaVA-3D
    & 3 & 16 & $2\times10^{-4}$ & Cosine
    & 3\% & 0 & BF16 \\
3D-LLaVA
    & 3 & 16 & $2\times10^{-4}$ & Cosine
    & 3\% & 0 & BF16 \\
LL3DA
    & 32 & 16 & $10^{-4}$ & Cosine
    & 1 epoch & 0.1 & FP32 \\
3D-LLM
    & 20 & 4 & $10^{-4}$ & Cosine
    & 1000 steps & 0.05 & AMP \\
PlaceIt3D-point
    & 20 & 16 & $10^{-4}$ & Cosine
    & 1000 steps & 0.05 & AMP \\
PlaceIt3D-region
    & 15 & 16 & $10^{-4}$ & Cosine
    & 1000 steps & 0.05 & AMP \\
\bottomrule
\end{tabular}
\caption{Fine-tuning hyperparameters from the launch scripts. Batch denotes effective batch size. Cosine schedules decay to $10^{-6}$ for LL3DA and $10^{-5}$ for 3D-LLM and PlaceIt3D; LL3DA warmup starts at $10^{-6}$.}
\label{tab:hparams}
\end{table*}

\paragraph{Hardware and inference.}
LLaVA-3D, 3D-LLM, and PlaceIt3D use one 80\,GB H100; 3D-LLaVA and LL3DA use one 40\,GB H100 partition. LLaVA-3D and 3D-LLaVA use greedy decoding, LL3DA uses beam search, 3D-LLM uses five beams, and PlaceIt3D uses placement-mask argmax. Seed 42 is specified for 3D-LLM and PlaceIt3D; the other scripts do not specify a seed.

\subsection{Instruction and Target Formats}
\label{app:prompts}

All models receive the same placement instructions without anchor coordinates. The formats below use \emph{put the yellow block to the left of the green ball} as an example.

\paragraph{Coordinate models.}
LLaVA-3D uses \texttt{<video>} followed by the instruction and predicts \texttt{[1.234, -0.567, 0.350]}. 3D-LLaVA uses \texttt{<image>} and predicts \texttt{1.23 -0.57 0.35}. Both use \texttt{human:}/\texttt{gpt:} roles. 3D-LLM stores the instruction in a \texttt{question} field and the bracketed coordinate string in an \texttt{answers} list.

\paragraph{LL3DA.}
LL3DA lowercases instructions and uniformly samples three training templates. Evaluation uses:

\begin{quote}
\ttfamily\small
\#\#\# human: given the 3D scene, \{instruction\} \#\#\# assistant:

the object should be placed at <loc>i, j, k</loc>
\end{quote}

The other training templates use \texttt{\#\#\# human: \{instruction\} \#\#\# assistant:} with target \texttt{<loc>i, j, k</loc>}, or prepend \texttt{in the given 3D scene,} with target \texttt{the target location is <loc>i, j, k</loc>}. Location integers are quantized using the scene's point-cloud bounds.

\paragraph{PlaceIt3D.}
Both variants use the unmodified instruction. PlaceIt3D-point queries its placement mask using \texttt{[SEG]}, while PlaceIt3D-region uses the mean question-token state. Both predict rotation buckets (\texttt{[ROT]}) and anchor masks (\texttt{[ANC]}).

\section{Expression Rephrasing Details}
\label{app:expression-naturalization}

This appendix provides the full prompts and implementation details for the candidate generation and verification + selection stages described in Section~\ref{sec:expression-naturalization}

\subsection{Candidate Generation Settings}
\label{app:candidate-generation-settings}
Qwen3-8B generates four candidate rewrites in a single inference call. We initially use deterministic decoding with temperature 0 and disable thinking mode to improve reproducibility and reduce unintended semantic variation. The maximum generation length is set to 512 new tokens. The model is instructed to return only a JSON object containing exactly four rewrites. Outputs that fail validation are regenerated with higher temperatures to encourage diversity. 


\begin{promptbox}{Generation System Prompt}
You are a precise natural-language rewriting assistant for spatial placement instructions.

Your task is to rewrite a template-like placement expression into multiple concise, fluent, human-like instructions while preserving its exact symbolic meaning.

The symbolic relations are the ground truth. The template expression supplies the intended object descriptions and instruction style.

SEMANTIC REQUIREMENTS

1. Preserve every distinct object appearing in the symbolic relations.
2. Preserve every spatial relation exactly.
3. Preserve the direction and roles of every relation:
   - figure is the object being located relative to the anchor.
   - anchor is the reference object.
   - For between, preserve both anchors.
4. Preserve all identifying attributes explicitly present in the template expression, such as color, size, material, or shape.
5. Do not infer or verbalize attributes merely because they appear inside a symbolic object identifier.
6. Keep multi-step relation chains attached to the correct objects.
7. Produce a placement instruction, not a scene description.
8. Use exactly one main placement action, such as "place," "put," "move," or "position."
9. Mention the placement target only once whenever a clear grammatical construction allows it.

RELATION MEANINGS

- left_of(A, B): A must be to the left of B.
- right_of(A, B): A must be to the right of B.
- in_front_of(A, B): A must be in front of B.
- behind(A, B): A must be behind B.
- between(A, B, C): A must be between B and C.
- nearest_to(A, B): A must be the nearest or closest object to B.
- farthest_from(A, B): A must be the farthest object from B.

IMPORTANT DISTINCTIONS

- "nearest_to" must remain superlative. Do not weaken it to "near," "beside," or "next to."
- "farthest_from" must remain superlative. Do not weaken it to "far from" or "far away from."
- "left of" and "right of" are not interchangeable.
- "in front of" and "behind" are not interchangeable.
- Do not reverse figure and anchor.
- A relation must not be omitted merely because another relation might imply it geometrically.

NATURALNESS GUIDELINES

- Use conventional spatial phrases.
- Remove mechanical or template-like wording.
- Avoid unnecessary repetition.
- Connect multiple relations naturally with relative clauses, participial phrases, or other unambiguous constructions.
- Prefer concise, single-sentence instructions.
- Vary sentence structure and placement verbs across rewrites.
- Make each rewrite meaningfully different, not just a punctuation change.
- Ensure modifiers and relative clauses clearly attach to the intended object.

PROHIBITED CHANGES

- Do not add, remove, merge, or substitute objects.
- Do not add object attributes.
- Do not remove attributes stated in the template.
- Do not add actions such as rotating, touching, aligning, picking up, stacking, or facing.
- Do not introduce "and then," "make sure," or "ensure" as a second action.
- Do not add, remove, reverse, weaken, strengthen, or infer spatial relations.
- Do not add spatial relations not explicitly present in the symbolic input.
- Do not refer to the image, scene, symbols, predicates, or rewriting process.
- Do not output explanations or reasoning.

PAIR ID REQUIREMENT

- Copy the input "pair_id" into the output exactly as provided.
- Do not omit, modify, regenerate, or renumber the pair ID.
- The pair ID must be a top-level field in the output JSON.

Before answering, silently perform these steps:

1. Build a ledger of all indexed objects.
2. Translate each symbolic relation into an exact directed fact.
3. Draft the requested rewrites.
4. Check every rewrite against the ledger:
   - all and only the required objects are present;
   - all relations are explicitly preserved;
   - direction and attachment are correct;
   - no attributes, actions, or relations were added;
   - nearest/farthest relations remain superlative.
5. Confirm that the output pair ID exactly matches the input pair ID.
6. Discard and replace any rewrite that fails a check.

Return valid JSON only, using exactly this schema:

{
  "pair_id": "<exact pair_id from the input>",
  "rewrites": [
    "rewrite 1",
    "rewrite 2",
    "rewrite 3",
    "rewrite 4"
  ]
}
\end{promptbox}

\begin{promptbox}{Generation User Prompt}
Generate exactly {NUM_REWRITES} distinct rewrites.

INPUT
{
  "pair_id": {PAIR_ID_JSON_STRING},
  "template_expression": {EXPRESSION_JSON_STRING},
  "relations": {RELATIONS_JSON}
}

Remember:
- Treat "relations" as the semantic ground truth.
- Use the template expression for the permitted object descriptions.
- Copy "pair_id" exactly from the input into the top-level "pair_id" field of the output.
- Output valid JSON only.
\end{promptbox}

\subsection{Candidate Verification and Selection Settings}
\label{app:verification-selection-settings}

We process the full dataset with the Gemini 3.5 Flash Batch API and store both the complete verification results and a compact JSONL file containing each
\texttt{pair\_id} and its selected naturalized expression. 

\begin{promptbox}{Verification + Selection System Prompt}
You are a strict semantic verifier and naturalness judge for spatial placement 
instructions.

For each record, you will receive:

- an original template expression;
- gold symbolic relations;
- four candidate rewrites.

Evaluate the candidates in two stages.

STAGE 1: SEMANTIC VERIFICATION

Judge each candidate independently.

A candidate is semantically valid only if it:

1. preserves the placement target;
2. preserves every anchor or reference object;
3. preserves every spatial relation, including its direction and argument roles;
4. keeps distinct indexed objects as distinct entities;
5. preserves both anchors in between relations;
6. preserves multi-step relation chains and modifier attachment;
7. preserves all identifying attributes stated in the original expression;
8. remains a spatial placement instruction with the same intended function;
9. does not add unsupported objects, attributes, actions, or spatial relations.

Use the gold symbolic relations as the ground truth for the required relations and argument roles. Use the original expression as the ground truth for explicitly stated object attributes and instructional intent.

Do not infer additional relations from geometry, common sense, or object identifiers.

For every invalid candidate, assign one or more of these error labels:

- `WRONG_TARGET`
- `MISSING_OR_CHANGED_OBJECT`
- `MISSING_RELATION`
- `ADDED_RELATION`
- `REVERSED_RELATION`
- `ARGUMENT_ROLE_ERROR`
- `BETWEEN_RELATION_ERROR`
- `RELATION_CHAIN_ERROR`
- `MODIFIER_ATTACHMENT_ERROR`
- `ATTRIBUTE_ERROR`
- `CHANGED_INSTRUCTION_INTENT`
- `UNSUPPORTED_ADDITION`
- `OTHER_SEMANTIC_ERROR`

Exclude every semantically invalid candidate from Stage 2.

STAGE 2: NATURALNESS VERIFICATION

Evaluate only the semantically valid candidates.

Assign integer scores from 1 to 3 for:

1. Grammatical correctness

- 1: serious grammatical problems
- 2: understandable but contains noticeable grammatical issues
- 3: grammatically correct

2. Fluency and natural wording

- 1: highly awkward, mechanical, or unidiomatic
- 2: acceptable but somewhat awkward or template-like
- 3: fluent, natural, and idiomatic

3. Clarity

- 1: difficult or ambiguous to interpret
- 2: mostly clear but contains slight ambiguity or awkward attachment
- 3: immediately clear, including its reference structure and modifier attachment

Do not reward lexical or syntactic diversity for its own sake. A candidate should not rank higher merely because it differs more from the original template.

Rank all semantically valid candidates from most natural to least natural.

First compare the three naturalness scores, using clarity as the first
tie-breaker, fluency and natural wording as the second tie-breaker, and
grammatical correctness as the third tie-breaker.

If two or more candidates still have identical scores, prefer, in order:

1. the candidate with clearer modifier attachment and reference structure;
2. the candidate with more idiomatic spatial wording;
3. the candidate that is more concise without omitting meaning;
4. the candidate with fewer unnecessary relative clauses or redundant words.

Do not change a candidate's scores merely to avoid ties.

If candidates remain genuinely indistinguishable after applying these
criteria, rank the lower candidate_id first.

Semantic correctness always takes priority over naturalness.

If no candidate is semantically valid:

- return an empty ranking;
- set `best_candidate_id` to `0`.

Return only JSON in the following format:

{
"candidates": [
{
"candidate_id": 1,
"semantically_valid": true,
"semantic_error_labels": [],
"grammatical_correctness": 3,
"fluency_and_natural_wording": 3,
"clarity": 3
}
],
"ranking": [1, 3, 2],
"best_candidate_id": 1
}

For invalid candidates, set all three naturalness scores to null.

The candidates array must contain exactly four entries with candidate_id values 1, 2, 3, and 4. Include every semantically valid candidate in the ranking, ordered from best to worst. Do not include invalid candidates in the ranking.
\end{promptbox}

\begin{promptbox}{Verification + Selection User Prompt}
Original expression:
{EXPRESSION}

Gold relations:
{RELATIONS_JSON}

Candidate rewrites:
1. {CANDIDATE_1}
2. {CANDIDATE_2}
3. {CANDIDATE_3}
4. {CANDIDATE_4}

For each candidate:

1. Determine whether it is semantically valid.
2. If valid, score it from 1 to 3 for:
   - grammatical_correctness
   - fluency_and_natural_wording
   - clarity
3. If invalid, do not assign naturalness scores.

Then rank all and only the valid candidates from most natural to least natural.

The first candidate ID in the ranking must be the best candidate.

If no candidate is valid:
- ranking must be []
- best_candidate_id must be 0

Return only the required structured JSON result.
\end{promptbox}

\section{Naturalness Evaluation Details}
\label{app:naturalness-evaluation-details}

\subsection{Annotators}

Two adult English-speaking evaluators based in the United States completed the human evaluation. One was an author with relevant research experience, and the other was an external volunteer recruited through the authors' personal network who had recently completed a bachelor's degree in computer science and had not participated in dataset construction. Both evaluators were informed of the purpose of the evaluation and consented to participate. The external evaluator received no financial compensation; participation was voluntary, and there was no employment, academic, or supervisory relationship between the evaluator and the authors.

\subsection{Sampling Procedure and Coverage}

We selected 100 expressions from the full dataset of 33,826 expressions for human evaluation. To ensure balanced coverage, we grouped expressions by placement type and selected the first six or seven examples from each group in their original dataset order.

For reference-structure and complexity coverage, the sample includes \textit{aligned} (7), \textit{between} (6), \textit{between\_one\_hop} (6), \textit{between\_two\_hop} (6), \textit{farthest} (6), \textit{multi\_relation} (6), \textit{nearest} (6), \textit{one\_hop} (6), \textit{one\_hop\_farthest} (6), \textit{one\_hop\_nearest} (7), \textit{ordinal} (7), \textit{pairwise} (7), \textit{table\_region} (6), \textit{two\_hop} (6), \textit{two\_hop\_farthest} (6), and \textit{two\_hop\_nearest} (6), covering all 16 placement types.


\subsection{Annotation Tasks and Instructions}
For each example, annotators were shown the original template expression, the gold symbolic relations, and four candidate rewrites. Each candidate was evaluated independently according to the following procedure.

\begin{appendixbox}{Stage 1: Semantic Validity}

Annotators first determined whether each candidate preserved the meaning of the original expression.

\begin{description}[
    leftmargin=1.6cm,
    labelwidth=1.2cm,
    labelsep=0.3cm,
    itemsep=3pt,
    topsep=3pt,
    font=\normalfont\bfseries
]
    \item[Valid] The target object, reference objects, spatial relations, and relation directions are all preserved.

    \item[Invalid] An important object or relation is added, removed, changed, reversed, or attached incorrectly.
\end{description}

A candidate was considered semantically valid only if it:

\begin{itemize}[leftmargin=*, itemsep=1pt, topsep=3pt]
    \item preserved the placement target;
    \item preserved all reference objects;
    \item preserved every spatial relation and its direction;
    \item preserved the roles of the target and anchors;
    \item kept indexed objects distinct;
    \item preserved both anchors in a \textit{between} relation;
    \item preserved relation chains and modifier attachment;
    \item preserved identifying attributes;
    \item remained a placement instruction; and
    \item did not introduce unsupported information.
\end{itemize}

The gold symbolic relations were treated as the ground truth for spatial relations and argument roles. Annotators were instructed not to infer additional relations from common sense or geometry.

If a candidate was invalid, annotators selected one or more of the following reasons:

\begin{itemize}[leftmargin=*, itemsep=1pt, topsep=3pt]
    \item Wrong target
    \item Missing or changed object
    \item Missing relation
    \item Added relation
    \item Reversed relation
    \item Target/anchor role error
    \item Between-relation error
    \item Relation-chain error
    \item Modifier-attachment error
    \item Missing or changed attribute
    \item Changed instruction intent
    \item Unsupported addition
    \item Other semantic error
\end{itemize}

Invalid candidates were not evaluated for naturalness.

\end{appendixbox}

\begin{appendixbox}{Stage 2: Naturalness}

For each semantically valid candidate, annotators assigned a score from 1 to 3 for each of the following dimensions.

\paragraph{Grammatical correctness.}

\begin{description}[
    leftmargin=0.7cm,
    labelwidth=0.3cm,
    labelsep=0.2cm,
    itemsep=1pt,
    topsep=3pt,
    font=\normalfont\bfseries
]
    \item[1] Noticeable grammatical problems.
    \item[2] Mostly correct, with minor issues.
    \item[3] Grammatically correct.
\end{description}

\paragraph{Fluency and natural wording.}

\begin{description}[
    leftmargin=0.7cm,
    labelwidth=0.3cm,
    labelsep=0.2cm,
    itemsep=1pt,
    topsep=3pt,
    font=\normalfont\bfseries
]
    \item[1] Awkward or unnatural.
    \item[2] Acceptable but somewhat awkward.
    \item[3] Fluent and natural.
\end{description}

\paragraph{Clarity.}

\begin{description}[
    leftmargin=0.7cm,
    labelwidth=0.3cm,
    labelsep=0.2cm,
    itemsep=1pt,
    topsep=3pt,
    font=\normalfont\bfseries
]
    \item[1] Difficult or ambiguous to interpret.
    \item[2] Mostly clear, with minor ambiguity.
    \item[3] Immediately clear and unambiguous.
\end{description}

Annotators were instructed not to reward a candidate merely because its wording differed more substantially from the original template expression.

\end{appendixbox}

\begin{appendixbox}{Final Selection}

Annotators ranked only the semantically valid candidates from most natural to least natural. They then answered:

\begin{center}
    \textit{Which candidate is the best overall rewrite?}
\end{center}

The available choices were:

\begin{itemize}[leftmargin=*, itemsep=1pt, topsep=3pt]
    \item Candidate A
    \item Candidate B
    \item Candidate C
    \item Candidate D
    \item None are semantically valid
\end{itemize}

\end{appendixbox}

\subsection{Annotation Agreement and Analysis}

Table~\ref{tab:human-gemini-agreement} reports the full agreement results for the human evaluation and Gemini comparison described in Section~\ref{sec:expression-naturalization}.

\begin{table}[h]
\centering
\small
\setlength{\tabcolsep}{3.5pt}
\renewcommand{\arraystretch}{0.95}
\begin{tabular}{@{}lrr@{}}
\toprule
Evaluation & Agreement & $\kappa$ \\
\midrule
Human semantic validity
    & 100.0\% & -- \\
Gemini-human semantic validity
    & 91.5\% & -- \\
Human best selection
    & 40.0\% & 0.19 \\
Gemini-human consensus
    & 62.5\% & 0.48 \\
\bottomrule
\end{tabular}
\caption{Human and Gemini evaluation results.}
\label{tab:human-gemini-agreement}
\end{table}

Both human annotators judged all 400 candidates to be semantically valid, while Gemini agreed with these judgments for 366 candidates (91.5\%). The 34 disagreements were concentrated in distance-based relations, where Gemini appeared to apply a stricter lexical interpretation than the dataset convention. For example, Gemini sometimes rejected expressions such as \textit{near}, \textit{next to}, or \textit{in proximity to} as realizations of \textit{nearest\_to}, and \textit{far away} or \textit{away} as realizations of \textit{farthest\_from}. These disagreements suggest that differences in how spatial relations are interpreted may affect LLM-based semantic verification.

The two human annotators selected the same best candidate for 40 of the 100 expressions (40.0\%, $\kappa = 0.19$). This relatively low agreement likely reflects the subjective nature of selecting among multiple semantically valid and similarly natural rewrites, rather than differences in semantic judgments. Among these 40 expressions, Gemini selected the same candidate as the human consensus in 25 cases (62.5\%, $\kappa = 0.48$). This suggests that Gemini's preferences align more closely with human judgments when the annotators themselves agree on the preferred rewrite.

\section{Placement-Location Evaluation Details}
\label{app:placement-location-eval}

\subsection{Annotators}

Three adult English-speaking annotators independently completed the placement-location evaluation. Each annotator evaluated the same 100 instructions and completed every item. All annotators received the same task instructions and used the same annotation
interface.

\subsection{Sampling Procedure and Coverage}

We selected 100 instructions from the full dataset of 33,826 instructions. Sampling was stratified to provide approximately balanced coverage across all 16 placement types. The \textit{pairwise}, \textit{one\_hop}, \textit{two\_hop}, and \textit{ordinal} types contributed seven examples each, and each of the remaining 12 types contributed six examples.

Grouped by relation family, the sample contains 28 directional examples and 18 examples each from the \textit{farthest}, \textit{between}, \textit{nearest}, and aligned/region/multi-relation groups. For the relation families with direct, one-hop, and two-hop variants, the sample contains 25 examples at each reference depth. The remaining 25 examples consist of \textit{ordinal} (7), \textit{aligned} (6), \textit{table\_region} (6), and \textit{multi\_relation} (6) instructions. Annotators were shown the rephrased version of each instruction.

\subsection{Annotation Tasks and Instructions}

For each example, annotators were shown the placement instruction, a rendered view of the scene, and an interactive 3D representation of the same scene. They were asked to select a location on the tabletop where the target object should be placed.

\begin{appendixbox}{Placement Selection}

Each annotator selected a location by clicking in the interactive 3D view. The interface displayed the target object at its actual size at the selected location so that annotators could assess whether it fit without colliding with another object. Placements that caused a collision could not be submitted, and
tabletop boundaries were enforced when placements were scored.

Annotators were allowed to rotate the 3D scene while completing the task. They were instructed to interpret directional terms such as \textit{left}, \textit{right}, \textit{in front of}, and \textit{behind} relative to the original camera view rather than the rotated view.

The interface also provided an optional automated relation check. Annotators were instructed to treat this only as feedback and to retain a placement if they believed that it correctly satisfied the instruction.

\end{appendixbox}

\subsection{Annotation Agreement and Analysis}

Two annotators produced valid placements for the same 97 of the 100 instructions. The third produced valid placements for 93. For the full sample, all three annotators produced valid placements for 93 instructions, all three produced placements that failed the evaluation criteria for three, and four instructions resulted in split judgments. So at least two of the three annotators produced a valid placement for 97\% of the sample. The pairwise agreement on the binary validity judgment ranged from 96\% to 100\%, with Cohen's $\kappa$ ranging from 0.58 to 1.00.

\begin{table}[h]
\centering
\small
\setlength{\tabcolsep}{4pt}
\begin{tabular}{lr}
\toprule
Outcome & Instructions \\
\midrule
All three valid & 93 \\
All three invalid & 3 \\
Split judgment & 4 \\
Majority valid & 97 \\
\bottomrule
\end{tabular}
\caption{Agreement on placement validity for the 100 human-evaluated instructions.}
\label{tab:placement-human-agreement}
\end{table}

The three unanimous failures consisted of two \textit{table\_region} corner placements and one \textit{multi\_relation} placement. Of the four split judgments, three were also \textit{table\_region} corner placements and one was an \textit{ordinal} directional instruction. Five of the seven instructions for which at least one annotator failed the evaluation were table-region examples. No failures or split judgments occurred for the \textit{nearest}, \textit{farthest}, or \textit{between} families, or for the one-hop and two-hop reference-chain conditions.

\section{Additional Results and Ablations}
\label{app:add-results}

\subsection{Zero-Shot Performance}
\label{app:zero-shot}

We evaluate zero-shot performance on templated instructions using the same validation examples and scorer as the fine-tuned models. LLaVA-3D achieves 14.5\% placement accuracy, with a 100\% parse rate, while the pretrained PlaceIt3D checkpoint achieves 2.5\% placement accuracy. The remaining coordinate-output models rarely produce usable placement outputs: 3D-LLaVA has a 0\% parse rate, LL3DA 1.0\%, and 3D-LLM 0.5\%. Their near-zero placement accuracy therefore reflects substantial output-format failure in addition to any limitations in spatial reasoning. Fine-tuning both teaches the task-specific output behavior and substantially improves relation and placement performance.

\begin{table}[h]
\centering
\small
\setlength{\tabcolsep}{4pt}
\begin{tabular}{lcc}
\toprule
Model & Placement acc. & Parse rate \\
\midrule
LLaVA-3D & 14.5 & 100.0 \\
3D-LLaVA & 0.0 & 0.0 \\
3D-LLM & 0.0 & 0.5 \\
LL3DA & 0.1 & 1.0 \\
PlaceIt3D (pretrained) & 2.5 & 100.0 \\
\bottomrule
\end{tabular}
\caption{Zero-shot results on templated instructions (\%). PlaceIt3D uses the authors' pretrained checkpoint.}
\label{tab:zero-shot}
\end{table}

\section{Reproducibility Details}
\label{app:repro}

\paragraph{Shared data and splits.} All models are built from the same rearrangement corpus and the same scene-level split assignment, so identical scenes appear in each split for every model. Rephrased instructions are merged by pair identifier. This guarantees that a given item has the same scene, target, anchors, relation, and goal under both phrasings. 

\paragraph{Scoring.} Metrics are computed by a single shared evaluator that is applied to all models. This is the reason that predictions with different native formats can be compared.

\paragraph{Computational budget and infrastructure.} All jobs used a single NVIDIA H100, either a full 80\,GB GPU (LLaVA-3D, 3D-LLM, and PlaceIt3D fine-tuning; 3D-LLM evaluation) or a \texttt{3g.40gb} MIG partition (3D-LLaVA and LL3DA fine-tuning; all other evaluations). We report wall-clock device-hours, counting each MIG partition as one device and using the latest completed run when multiple attempts exist. The twelve reported fine-tunes total $73.2$ GPU-hours: $10.5$ for LLaVA-3D, $6.6$ for 3D-LLaVA, $19.9$ for LL3DA, $21.0$ for 3D-LLM, $8.4$ for PlaceIt3D-point, and $6.9$ for PlaceIt3D-region, with each model trained on templated and naturalized instructions. The 23 reported evaluation jobs add $18.4$ GPU-hours, for $91.6$ GPU-hours across reported experiments. Including failed and superseded runs, LL3DA jobs reaching the 24-hour limit, feature lifting, and zero-shot evaluations, total project usage was approximately $203$ GPU-hours.

\subsection{Artifact Licenses}

This work builds on the CLEVR dataset and generation framework. CLEVR data are released under the Creative Commons Attribution 4.0 license and the CLEVR dataset-generation code is distributed under a BSD license. These licenses permit the adaptations and distribution used in this work, subject to their respective attribution and notice requirements.

\end{document}